# Lexicon generation for detecting fake news

Uğur Mertoğlu [1], Burkay Genç [2]

[1] Department of Computer Engineering, Hacettepe University, Ankara, Turkey; umertoglu@hacettepe.edu.tr
[2] Department of Computer Engineering, Hacettepe University, Ankara, Turkey; bgenc@cs.hacettepe.edu.tr

**Abstract:** With the digitization of media, an immense amount of news data has been generated by online sources, including mainstream media outlets as well as social networks. However, the ease of production and distribution resulted in circulation of fake news as well as credible, authentic news. The pervasive dissemination of fake news has extreme negative impacts on individuals and society. Therefore, fake news detection has recently become an emerging topic as an interdisciplinary research field that is attracting significant attention from many research disciplines, including social sciences and linguistics. In this study, we propose a method primarily based on lexicons including a scoring system to facilitate the detection of the fake news in Turkish. We contribute to the literature by collecting a novel, large scale, and credible dataset of Turkish news, and by constructing the first fake news detection lexicon for Turkish.

**1. Introduction**

Even though the preference to obtain news from traditional media sources like TV, press etc. is still not negligible, the consumption of digital media sources is rapidly increasing. People follow the news via the internet rather than traditional methods, notably through social media now. In 2019, about 2.82 billion users are estimated [1] to use social media worldwide and the number is projected to increase to almost 3.1 billion by 2021, that is nearly half of the world population. Providing global connectivity, social media platforms such as Facebook, Twitter and WhatsApp make fake news far more influential on society than the conventional news platforms do thanks to the domino-effect they cause.

In recent years there has been a lot of discussions on ``fake news'' about identifying and combating fake news. Because it is a great threat to democracy, economy, and journalism around the world as Zhou et al. stated [2]. Being exposed to ``fake news'' at an unprecedented pace, it appears most readers are not capable of questioning the credibility of these news and distinguishing whether they are fake or not.

Fake news has a huge potential to manipulate people's perception of reality and view of the world through growing mistrust among people. Moreover, it causes major problems for politics, tourism, national security, healthcare systems, society polarization (conflicts and violence among ethnic groups, refugees, immigrants) etc. For instance, causes behind the outbreak of the Arab Spring, which first emerged in Tunisia and spread across the Arab World in a short period of time, are numerous such as imbalances in income, corruption, widespread poverty etc. However, it was some fake news distributed in social media [3] that ignited the spread of the events by making the situation even more chaotic.

Due to the rapid changes in communication patterns, not only people are vulnerable to the bombardment of information and news they are exposed to but so are the governments and organizations. For example, social media platforms like Facebook, Twitter and Google etc. have started to work for solutions [4] to preserve their reputation. Moreover, in accordance with the growing academic studies about the subject, international panels, conferences, and many activities have been organized. However, it seems that state of the art systems of these efforts need to be enhanced because now the unsustainable struggle mostly depends on human review, crowd-sourcing, removing fake accounts, fact-checking organizations, promoting media literacy and third-party tools. With more than a billion pieces of content posted every day we know that fact-checkers cannot review every story manually. Therefore, looking into automated ways to act on a bigger scale using computer scientific methods has a reasonably strong motivation.

In this study, we present a new analysis method that is based on lexical resources derived from a large corpus. We utilized lexicon generation methodology for language modeling of fake and valid news texts. Lexicon-based approaches are widely used in the studies in many types of NLP analyses, especially when the goal of the study is to detect polarity [5-7] as we did in our work. However, to the best of our knowledge, there is no study using lexicons, language models or dictionaries which have fake-valid polarities of language components. One of the most important reasons for this absence is the non-existence of language specific models or lexicons in terms of fake news. To lead the field in this respect, we developed lexicons to be a basis for generating sophisticated fake news lexicons in Turkish. In this lexicons, we assigned Fake/Valid values for each of ``raw word'', ``raw word and its POS tags'', ``root/stem of word'' (root will be used throughout the study) and ``suffixes'' in Turkish language.

Throughout this study, we will use the term ``Valid'' to represent ``Non-fake'' news. Our aim is not to validate the content of news from a sociological or technological perspective. From our standing point, a news story on a politician claiming that ``the moon is green'' is valid news if the politician indeed made the claim. Hence, we deliberately do not use the terms ``True News'' or ``Correct News'' as that implies the correctness of the content. Rather we use ``Valid News'' to show that the story is ``**credible**'', or worth ``**considering**''. Our study contributes (a) a novel lexicon construction model to detect fake news in Turkish and (b) an adaptable workflow offering an insight into fake news detection for other languages especially agglutinative ones such as Finnish, Hungarian etc. if the structural diversity of the language is properly addressed. This work can also be seen as an exploratory data analysis of language for fake news detection before using machine learning or deep learning techniques.

The rest of the paper is structured as follows: Section 2, presents a review of the relevant literature. The data and methodology we use is discussed in Section 3. In Section 4, results and evaluations are presented. Finally, in Sections 5 and 6, we provide discussion, conclusions and propose possible future study topics.

## 2. Related Work

The increasing amount of fake news can be attributed to mankind's tendency to spread crowd-pulling information as reported in a study of Vosoughi et al. [8]. And one of the many ways to do this is textual deception which is now flagrant in online text. Natural language processing (NLP) and text mining studies in the literature divides textual deception into many sub-categories; that is, it relates textual deception to other closely related / adjacent text classification problems. To make this comprehensive literature survey more understandable, we visualized it in Figure 1. In terms of fake news detection, many systems have been proposed, utilizing several techniques those are reviewed in this section.

| Verbal Communication | Computer Mediated Communication | Law, Security and Science | Journalism |
|---|---|---|---|
| • Psychological/Clinical Cases | • Emails (Spam, phishing, bot etc.) | • Court Judgements | • Politic, Economic Interests |
| • Questionnaires, Interviews | • Forum, Blogs, Online Services | • Forensic Science (Police interrogations) | • Regional, Global Rivalry/Competition |
| • Case Scenarios | • Social Media (Fake accounts, tweets, posts, links etc.) | • Intelligence Reports | • Social Events (Revolts, Protests, Disturbance etc.) |
| • Customer Services Talks | • Reviews (Product, hotel, services etc.) | • Forgery on Documents | • Company/Brand Competition |
| • Empirical Data (Polygraph Tests) | • Advertising | • Suspect/Witness Reports | • Exploitation of Social Media/Internet |

**Figure 1.** Variations of textual deception in the literature.

Although the chronology may also be linked to earlier, following the 2016 US presidential election many have expressed concerns about the effects of fake news [9] and the topic has gained popularity lately. Gröndahl and Asokan [10] associated the popularity of the topic among researchers from Information Technology (IT) departments to inefficacy of non-expert human performance. This reason motivated researchers for finding computational solutions.

In one of the pioneering works on deception detection [11] Conroy et al. make a summarization and explain the existing methodologies related to the fake news detection. The listed methodologies mainly focus on linguistic approaches, network approaches, source credibility approaches, semantic approaches, and hybrid approaches. The range of studies include relatively simple linguistic approaches, such as ``bag of words'' as well as complicated deep learning methods. And researchers concentrated on various domains. Some of the domains used in these studies are detecting rumors in social media posts [12] detecting spam posting [13], gender deception in online communication [14], deceptive opinions in online reviews [15].

One of the most used approach is the linguistic approaches by the researchers. For instance, Markowitz and Hancock used linguistic approach to find clues of deception utilizing n-grams, part-of-speech (POS) tags and other syntactic analysis methods [16]. Some others tried to improve their results by using the linguistic approach as a complementary tool [17,18].

In the intersection of the semantic and linguistic approach, rhetorical-base detection is used by some studies. Rubin used rhetorical structure theory (RST) as the analytic framework to identify systematic differences between deceptive and truthful stories in terms of their coherence and structure in her study [19]. Accordingly, in a study focused on discourse level, rhetorical structures are used as vector space modelling applicants for predicting whether a report is truthful or deceptive for English news [20].

Some researchers use traditional machine learning methodology pointing to drawbacks of deep learning techniques about stance detection [21] whereas some researchers have applied deep learning techniques reporting shortcomings of more traditional machine learning techniques. In one of the works [22] using deep learning, the authors have developed models based on a pre-trained convolutional neural network (CNN) for extracting sentiment, emotion and personality features for sarcasm detection which is sub-domain of fake news.

In almost all the studies, researchers used publicly available datasets in English, given the fact that collecting and labelling such data is a time-consuming process. Moreover, these studies generally focus on fake news detection via machine learning and deep learning techniques. It is known that relatively more successful results can be obtained by studying a specific domain in machine learning problems. Therefore, these studies mostly focus on a specific domain such as politics or sports news by regarding as a binary classification problem.

In this work, we focus on textual deception in Turkish news by developing lexicons. We presented to what extent the fake news can be detected without machine learning techniques and to reveal the tacit knowledge used to deceive readers in the text.

**3. Materials and Methods**

To construct a robust, general purpose fake news lexicon in Turkish, we focused on including many types of fake news. We used Zemberek [23], the Turkish NLP engine, to obtain roots of words, POS tags and suffixes. We developed the first Fake News Lexicon Model for Turkish, named FaNLexTR containing 4 different categorizations derived from a comprehensive corpus. We propose a novel methodology to construct general purpose fake news lexicons for Turkish.

We generate the lexicons utilizing a large labeled corpus of news texts using the GDELT (Global Database of Events, Language and Tone) Project [24] datasets, verified data taken from a fact-checking organization ``teyit.org'' and online news data manually verified by our research team.

*3.1. Data Collection and Preparation*

In agglutinative languages such as Turkish, words are extended by suffixes to create new lexemes. Therefore, i.e. a lemma, can be extended in tens of different ways to obtain many different lexemes. We concentrated on gathering a large body of news texts which we can use as a training set for FaNLexTR and other possible future research studies. While creating our initial database of news texts, the validation and accuracy of data labelling was one of our highest priorities. To this end, we used the archives of the GDELT project to obtain the URLs of approximately 100k news published by 3 major, authentic news agencies (DHA, IHA, AA) in Turkey. Hence, our news texts database does not include any news from local or hard to validate news sources. The fake news part of the database is constructed by articles from ``teyit.org'', a fact-checking organization in Turkey, which tags the news as Fake or True, and publishes them on its web page. Furthermore, we include a hand curated collection of fake and valid news obtained from various online sources manually verified by our research team. We refer to this collection as Manually Verified News, MVN. These three methods that we prefer while collecting data are significant in terms of representing methods for verification of news in real world. These methods consist of artificial intelligence projects and tools, fact-checking organizations, and human effort, respectively.

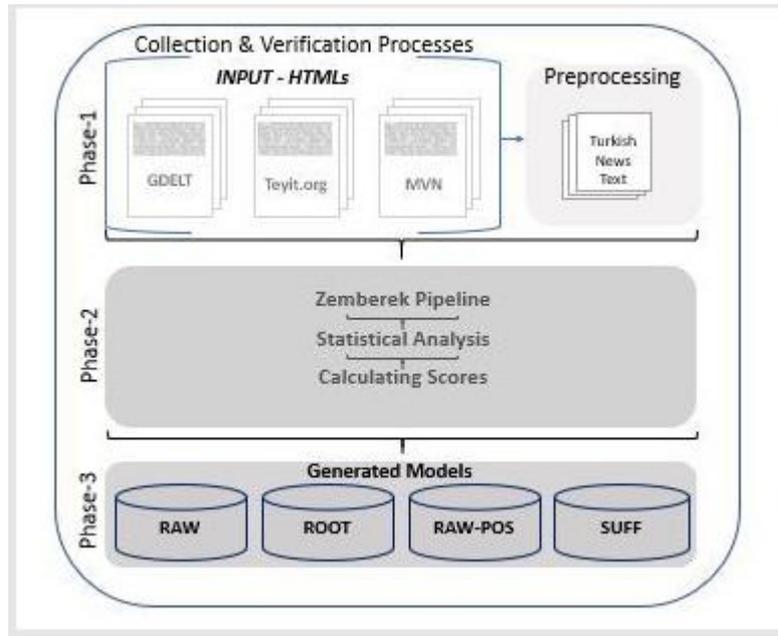

**Figure 2.** Main phases of FaNLexTR development framework.

Our dataset collection steps, and main phases of the process are outlined in Figure 2. In the first phase, we scraped the data from the web, cleaned HTML tags and other extra materials, obtained lean news content, and stored them in files. The final dataset consists of 84734 unique news articles, which belong to 11 different news domains, uniformly distributed within a period between years 2016 (last 2 months) and 2019. The corpus statistics, including class distribution and word/sentence statistics, are shown in Table 1. In the second phase, we ran all news texts through the Zemberek pipeline to obtain the unique words in each text, as well as the associated information for each word, such as root of the word, part of speech information and the ordered set of suffixes.

As seen from this table, the fake news count is quite small compared to valid ones. It may seem like the imbalanced nature of the dataset can cause a handicap, however this is not an unexpected issue, as real world datasets are mostly composed of ``normal'' observations with only a small percentage of ``abnormal'' or ``interesting'' examples. It should also be noted that ``falsehood diffuses significantly farther, faster, deeper, and wider than the truth in all categories'' [8]. Moreover, we do not use any classical machine learning models, some of which are highly sensitive to the imbalance of training sets. Furthermore, the average token and sentence count statistics are much smaller in fake news when compared with valid news. This is an expected phenomenon as much simpler language is mostly used in fake news whereas in valid news sophisticated (lexical richness) expressions are used. From now on, we refer to the valid news training dataset as $D_{tr}^{V}$, valid news test dataset as $D_{ts}^{V}$; fake news training dataset as $D_{tr}^{F}$, and fake news test dataset as $D_{ts}^{F}$.

**Table 1.** The collected news data statistics: type, source and class of the dataset, number of news texts within the dataset.

| Dataset | Source | Class | Count |
|---------|--------|-------|-------|
| Train | GDELT | VALID | 82691 |
| Train | MVN | VALID | 855 |
| Train | Teyit.org | FAKE | 902 |
| Train | MVN | FAKE | 286 |
| Test | GDELT | VALID | 22 |
| Test | MVN | VALID | 188 |
| Test | Teyit.org | FAKE | 89 |
| Test | MVN | FAKE | 121 |

*3.2. Data Verification*

The most challenging and time-consuming process of this kind of work is data collection and preparation. Because representing the real-world needs meticulously effort of answering the question when a news is valid or fabricated. To provide this, data validation and data verification process must be executed in correct manner.

In our collection, there are two sets already validated by GDELT and teyit.org. And our team validated the MVN part of our dataset. The news collected through GDELT comes from three respective sources only. However, this does not mean they never ever serve fake news. So, we made cross-checks between our data sources to exclude any mislabeling. In other words, we checked whether the news obtained from GDELT included any news that are already in our fake news datasets. Considering that these are mainstream news agencies, any potential fake news content delivered by them is supposed to be an immediate attraction for fact-checking organizations such as teyit.org. For the verification, we applied some computational and statistical analysis. On the assumption that certain slang words are more likely to be used in fake news. We used the slang dictionary [25] for Turkish. And we compiled a list of 782 Turkish slang words and phrases, and we compute the mean of slang words occurrence per sentences for in the datasets. And we also did the similar experiment for spelling errors considering this can be a sign of sloppy journalism. We saw highly correlated results with the notion that can be attributed to the correctness of labeling shown in Table 2.

**Table 2.** Means of slang words occurrence per sentences and spelling errors per sentences.

| Dataset | Slang words occurrence per sentences | Misspelling of words per sentences |
|---------|--------------------------------------|------------------------------------|
| GDELT (Non-fake) | 0.032 | 0,057 |
| MVN (Non-fake) | 0,071 | 0,102 |
| GDELT (Fake) | 0,068 | 0,091 |
| Teyit.org (Fake) | 0,115 | 0,396 |
| MVN (Fake) | 0,098 | 0,275 |

These statistics give us clear signs of formal and informal language usage which show their tendency to sophisticated or sloppy journalism. As we beforementioned in Section 1, we do not try to verify the content, we are trying to predict Valid News and Fake News (Fabricated News).

*3.3. Methodology*

Turkish is an agglutinative language where suffixes are added to the end of a word to change its meaning and use in a sentence. It is important to explain the main components here briefly as follows:

A lemma stands for the canonical form of an inflected word, i.e. the form usually found in dictionaries (e.g. run). And a lexeme stands for all the inflected forms of a word (e.g. run, runs, running, ran, runner, etc.).

Hence, we have the following structure in a Turkish word:

$$Lexeme = Lemma + S1 + S2 + \cdots + Sk$$

Here, Lexeme is the raw form used in text, Lemma is the root/stem of the word, and $S_i$'s are the suffixes. Although, the root word *Lemma*, generally determines a major part of the meaning of *Lexeme* certain suffixes can have a significant effect on the meaning. Indeed, even the part of speech information of the word may be determined by suffixes. Hence, we decided to split our analysis in four different classes of information: RAW words, ROOTs, RAW+POS tag, and SUFFIXes. For each of these classes, we aim to generate separate models and compare them at the end to see which class model provides the best classifier for validity of Turkish news.

The main idea behind our approach is as follows: we first compute the frequency score of each term (raw, root, raw+pos or suffix) in $D_{tr}^V$, we then compute the corresponding frequency scores for $D_{tr}^F$, and finally we compute a novel metric named Fake/Valid score for each document to be tested by summing up the Fake/Valid frequencies of the terms contained within the document.

3.3.1. Lexicon Generation

The first thing we need for constructing lexicons based on the above four classes is to extract the corresponding terms of that class from the news datasets. To this end, we process the $D_{tr}^V$ and $D_{tr}^F$ datasets and extract the full lists of RAW, ROOT, RAW+POS and SUFFIX terms within the documents. For each term in each list, we compute two scores: one based on its frequency in $D_{tr}^V$, and another based on its frequency in $D_{tr}^F$. The scores $S_{t,C}^V$ and $S_{t,C}^F$ for term *t* and model class *C* are formally represented in Equations (1) and (2).

$$S_{t,C}^V = \frac{\sum_{d \in D_{tr}^V} f_d(t)}{\sum_{d \in D_{tr}^V} \sum_{x \in T_C} f_d(x)} \quad (1)$$

$$S_{t,C}^F = \frac{\sum_{d \in D_{tr}^F} f_d(t)}{\sum_{d \in D_{tr}^F} \sum_{x \in T_C} f_d(x)} \quad (2)$$

In these equations, $T_C$ represents the set of all terms within the model of class *C*, where *C* is one of RAW, ROOT, RAW+POS or SUFFIX. Also, $f_d(x)$ represents the frequency of term *x* in document *d*. After computing the fake and valid term scores for all four models, we obtained the resulting statistics shown in Table 3. Note that each model now contains terms from both $D_{tr}^V$ and $D_{tr}^F$ datasets. We now explain these results briefly.

**Table 3.** The models developed.

| Model | Unique Term | Common Term | Only in Fake | Only in Valid |
|---|---|---|---|---|
| RAW | 443174 | 10166 | 456 | 432552 |
| ROOT/STEM | 63237 | 3830 | 77 | 59330 |
| RAW+POS | 457187 | 10301 | 466 | 446420 |
| SUFFIX | 9930 | 898 | 12 | 9020 |

The RAW Form Model is the most intuitive model where we only consider the default form of the words as they occur within the texts. Hence, there is a high number of unique words in this model. However, when we consider words common to both valid and fake news texts, the number falls to 10166 unique words. This is the result of two mechanisms in action: first, $D_{tr}^{F}$ is much smaller than $D_{tr}^{V}$ and hence contains much less words; second, as we have mentioned earlier the language used in fake news is much simpler, resulting in less unique words. On the other hand, there are 456 unique words which only exist in the fake news texts. When we examine these words, we see that these are mostly informal words and exclamations which cannot be used in formal, valid news texts. The existence of these words in a text is also a very strong indicator of the fakeness of the text. Some interesting term examples and their associated fake and valid scores in RAW Form Model are provided in Table 4.

**Table 4.** Some examples from the RAW Form Model. The terms on the left have greater valid scores, and the terms on the right have greater fake scores. All scores are multiplied by 10000 for readability.

| Raw Word | Fake score | Valid score | Raw Word | Fake score | Valid score |
| --- | --- | --- | --- | --- | --- |
| Ardından | 14,80 | 21,80 | Zaten | 60,62 | 18,60 |
| Bulunan | 7,85 | 17,09 | Bulunmuş | 0,6 | 0,1 |
| İlk | 10,29 | 20,12 | Son | 17,21 | 15,59 |
| Tutuklanarak | 0 | 1,19 | Serbest | 3,62 | 2,95 |
| Yok | 5,74 | 6,22 | Değil | 11,79 | 7,07 |
| Doğrultusunda | 0 | 1,59 | Gibi | 32,32 | 17,60 |
| Korktu | 0 | 0,01 | Tırstı | 0,3 | 0 |
| Güvenilmez | 0 | 0,05 | Kaypak | 0,6 | 0 |
| Mantıksız | 0 | 0,01 | Saçma | 1,20 | 0,06 |

**Table 5.** Some examples from the ROOT Form Model. The terms on the left have greater valid scores, and the terms on the right have greater fake scores. All scores are multiplied by 10000 for readability.

| Root/Stem | Fake score | Valid score | Root/Stem | Fake score | Valid score |
| --- | --- | --- | --- | --- | --- |
| Süs | 0 | 0,29 | Janjan | 1,21 | 0 |
| Ara | 28,70 | 38,90 | Bul | 8,15 | 7,26 |
| Değer | 6,64 | 13,37 | Haber | 17,82 | 7,49 |
| Sahtekâr | 0 | 0,03 | Sahte | 1,81 | 1,56 |
| Hilekâr | 0 | 0,004 | Fırıldak | 2,41 | 0,001 |
| Net | 0,91 | 1,55 | Gibi | 32,31 | 17,86 |
| Düzgün | 0 | 0,29 | Paçoz | 0,3 | 0 |
| Yüzde | 0 | 4,07 | Tahmini | 1,21 | 0,03 |

When we only consider the roots/stems of the words, the number of unique terms drops down to 63237. The number of common terms in fake and valid texts are again much less than the total number of terms. We have detected 77 root terms that are unique to fake news. These are mostly slang or made-up words that are not part of the formal language. Again, these are strong indicators of fake news. Selected term examples and their associated fake and valid scores in ROOT Form Model are provided in Table 5.

The RAW+POS Form Model uses the raw forms of the words paired with their part-of-speech tags. Therefore, the same word can exist multiple times in this model, each time paired with a different POS tag. We can see from Table 3, that there are 457187 unique word-POS pairs, of which 466 is unique to fake news and 10301 are common. Selected term examples and their associated fake and valid scores in RAW+POS Form Model are provided in Table 6.

**Table 6.** Some examples from the RAW+POS Form Model. The terms on the left have greater valid scores, and the terms on the right have greater fake scores. All scores are multiplied by 10000 for readability.

| Raw + Pos | Fake score | Valid score | Raw + Pos | Fake score | Valid score |
|---|---|---|---|---|---|
| Teröristlerce | 0 | 0,27 | Terörist | 3,32 | 2,11 |
| Apaçık (Sıfat) | 0 | 0,02 | Alenileşen | 1,21 | 2,33 |
| Bitmemiş (Fiil) | 0 | 0,01 | Bitmiyormuş | 0,3 | 0 |
| Sımsıkı (Zarf) | 0 | 0,001 | Yapışırcasına | 0,3 | 0 |
| Uyanık (İsim) | 0 | 0,07 | Uyanık | 1,21 | 0,1 |
| Olay (İsim) | 6,04 | 15,59 | Olaylara | 0,6 | 0,3 |
| Biz (Zamir) | 11,17 | 12,33 | Gibi | 1,81 | 0,1 |
| Sınır (İsim) | 1,21 | 1,38 | Hadleri | 0,6 | 0 |
| Olduğunu | 18,72 | 22,70 | Olarak | 44,70 | 41,04 |

The last model is based on the suffixes. This is probably the most interesting model we have constructed, considering that no suffix combination lexicons exist in the literature and this can only be done for agglutinative languages. In the suffix lexicon, we considered all possible sub-sequences of the suffixes of a word. For example, if $R + S_1 + S_2 + S_3$ is a word that exists in RAW Form Model, then in SUFFIX Form Model we consider $S_1$, $S_2$, $S_3$, $S_1+S_2$, $S_2+S_3$ and $S_1+S_2+S_3$ as terms. Overall, there were 9930 detected suffix sequences, 898 were common in both news types and only 12 belonged exclusively to fake news. Selected term examples and their associated fake and valid scores in SUFFIX Form Model are provided in Table 7.

**Table 7.** Some examples from the SUFFIX Form Model. All scores are multiplied by 10000 for readability.

| (SUFFIXes) | Fake score | Valid score | Word |
|---|---|---|---|
| Caus-Caus-Neg-FutPart-A3pl-Acc | 4,11 | 0 | ARA -t-tır-ma-yacak-lar-ı |
| Inf2-P3pl-Narr | 1,54 | 0 | AT -ma-ları-ymış |
| A3pl-Loc-Rel-P2sg-Abl | 3,86 | 0 | ZAMAN -lar-da-ki-n-den |
| PresPart-P3sg-Narr | 1,54 | 0 | GİD -en-i-ymiş |
| A3pl-Dat | 111,91 | 9,49 | İNSAN -lar-a |
| With-A3pl-P2sg | 0 | 5,8 | TALİH – li-ler-in |
| Able-Aor-A1pl | 0 | 5,00 | KAÇ -abil-ir-iz |

So far, the construction of four models has been explained. Next, we show how to use these models to evaluate the validity/fakeness of a document.

3.3.2. Document Evaluation

In this part, we explain and demonstrate how to use the generated models to evaluate the validity/fakeness of a given document. First, we need to note that considering we have generated 4 different models based on different terms, we will be conducting 4 different analyses. The text we will use for demonstration purposes is provided in Table 8. In summary, the text boasts about Cuba as a holiday destination and reports a few statistics about the country.

**Table 8.** An example fake news text from our testing dataset.

| **News Header** | İNANILMAZ AMA DOĞRU |
|---|---|
| **News Text** | Ta Küba! Kim gidecek demeyin! Heralde bu yaz tatil listenizdeki yer Küba olmalı. 47 yıldır cinayet işlenmedi. 58 yıldır tecavüz ve istismar suçu işlenmedi. Hatta 5 yıldır hırsızlık bile olmadı. Herkese eşit maaş. Vergi yok. Hemen herşey ücretsiz ya da sudan ucuz. Gezin görün! |

To analyze and evaluate a document we execute the following steps: first, we parse the text and extract the related terms for each model, then we sum up the corresponding fake and valid scores of terms over all text. The resulting sum of fake scores of terms is called the document fake score, $S_{D,C}^F$, and the sum of valid scores of terms is called the document valid score, $S_{D,C}^V$. The formal definitions of both are provided below:

$$S_{D,C}^V = \sum_{t \in T_C^D} S_{t,C}^V \quad (3)$$

$$S_{D,C}^F = \sum_{t \in T_C^D} S_{t,C}^F \quad (4)$$

In Eq. 3, $T_C^D$, represent the set of all terms in document *D* with respect to model class *C*. Once these two scores are computed for each model class, we compare to see which one is greater. If $S_{D,C}^V$ is greater than $S_{D,C}^F$, then we label the document as VALID with respect to model *C*, otherwise we label it FAKE:

$$L_D = \begin{cases} VALID, & if\ S_{D,C}^V > S_{D,C}^F \\ FAKE, & else \end{cases} \quad (5)$$

Let us now look at the example news text in Table 8. This example demonstrates a click-bait, a text that is generated to provoke clicks. Most of the time, click-baits are hidden between valid news texts and contain false information. This text is one of the shortest in our test dataset. The reason we have chosen it is to be able to fit the complete analysis into these pages. An analysis conducted on a larger text follows the same steps as this one.

We will start by the RAW Form Model. With respect to this model, the text contains 39 different terms. Of these terms we want to mention a few that have striking differences between $S_{t,C}^F$ and $S_{t,C}^V$. Let us start with ``heralde'', which means ``in any case''. However, there is an important issue here: the correct spelling of this term is ``herhalde''. However, the middle `h' in this word is a very weak `h' and in daily speech mostly it is not pronounced. Still, a valid and respected news source should use the correct form of the word. We see from the RAW Form Model that this word (``heralde``) is indeed occurring in both valid and fake news, however it is almost 600 times more frequently used in fake news. The difference between $S_{t,C}^F$ and $S_{t,C}^V$ in this case is 0.603. Another interesting word is ``bile'' which means ``even'' in English, as used in the sentence ``There wasn't even a theft in 5 years.''. We can see from the model scores that ``bile'' is a very frequently used function word, both in valid and fake news. However, the difference between $S_{t,C}^F$ and $S_{t,C}^V$ is very large, making use of ``bile'' a strong indicator of fake news. A few more interesting terms are provided in Table 9.

**Table 9.** Some example RAW Form Model terms from the example news text in Table 8. All scores are multiplied by 10000 and rounded down for a nicer presentation.

| Term | $S_{t,C}^F$ | $S_{t,C}^V$ |
| --- | --- | --- |
| heralde | 0,604 | 0,001 |
| listenizdeki | 0,302 | 0,0003 |
| herşey | 0,604 | 0,0286 |
| görün | 0,906 | 0,0418 |
| tecavüz | 2,7181 | 0,281 |
| Bile | 8,1544 | 2,5641 |

Overall, the $S_{D,C}^V$, $S_{D,C}^F$ scores are computed as 50.20 and 70.37 (scores multiplied by 10000 for better readability). This is a clear win for $S_{D,C}^V$ and hence the RAW Form Model labels the text as FAKE.

When we consider the ROOT Form Model, we again detect 39 root terms. However, there is a critical issue here. Extracting roots from Turkish words is not a simple feat, and Zemberek, the Turkish NLP engine we use is not perfect. Therefore, we notice some incorrect terms when we manually examine the results. However, to keep things as automated as possible, we do not fix these mistakes. Once again ``heralde'' tops the list of interesting terms as it is also detected as a root term. The overall $S_{D,C}^F$ and $S_{D,C}^V$ scores are 504.06 and 493.46, respectively. It is a closer call, but still a win for $S_{D,C}^F$.

Next, we look at the RAW+POS Form Model which pairs raw words with their POS tags. Although this creates some variety, the overall scores are similar to the RAW Form Model: 62.517 for $S_{D,C}^F$ and 44.634 for $S_{D,C}^V$ Once again a decisive win for $S_{D,C}^F$.

The last model is the SUFFIX Form Model. As we have mentioned earlier, this model only uses the suffix morphemes of the words, completely ignoring the root. In this news text we observed 25 different terms (suffix sets). One of the interesting suffixes to note is the A2pl (second person plural) suffix. This suffix has a $S_{t,C}^F$ of 16.980 and a $S_{t,C}^V$ of 4.077. This shows that, despite being used in valid news, the use of A2pl is much more frequent in fake news. Naturally, one does not expect a news text to be written with direct references to the reader. However, in this example the word ``demeyin'', which literally means ``do not say'', is directed at the reader as if the author of the text is speaking to the reader. The use of this suffix is significantly penalized by our model. As a result, $S_{D,C}^F$ and $S_{D,C}^V$ become 981.9913 and 954.4907, respectively. Once again, the document is classified as FAKE.

At this point, we want to discuss two challenging properties of the Turkish language. The first one is the morphological ambiguity problem, which means a word in its raw form may have multiple ambiguous morphological analyses. For example, the word ``telaşına'' in the sentence ``Telaşına geldi herhalde, anahtarı unuttun.'' may be analyzed either as Noun+A3sg+P3sg+Dat or as Noun+A2sg+P2sg+Dat. Zemberek, the NLP engine we are using in this study is suggesting the first one, however the correct analysis with respect to the meaning within the sentence is the second one. To choose the right analysis, one must look at the verb "unuttun". The verb hints that the object of the sentence is "you" and not "he" or "she". Therefore, the right analysis with respect to the object becomes the second analysis. Unfortunately, this analysis requires a significant effort to be automated, and is not our focus in this study. This ambiguity mostly affects the SUFFIX and RAW+POS models and even then, the effect is negligible. The reason for this claim is that we are following a bag of words approach in evaluating the documents, and wrongly analyzing one or two words in a document does not create a significant deviation in the document score. However, we are aware of the possible improvement opportunity available here and hope to deepen our analysis in our future studies.

The second challenging property we want to discuss is the sparseness problem. Being an agglutinative language, in Turkish words are constructed by appending multiple suffixes to a root. In conclusion, many root-suffix combinations may have zero frequency and full listing hypothesis cannot be applicable [26]. However, not all suffix combinations make sense, and not all combinations that make sense are applicable to all words. This means, even if you consider all possible words in the Turkish language, some suffix combinations will never appear, and some will only appear very rarely. In the works of Arısoy et al. [27], sub-lexical units are offered to deal with this problem in Turkish. In another study [28] the authors use character and morpheme information.

Although this may look like a problem for our models, in fact they are using it to their benefit. Whenever a suffix combination appears very rarely in the language, it tends to appear only in fake news or valid news. This makes the suffix combination a strong indicator of the fakeness/validity of the document, hence its score increases significantly.

Considering that the agglutinative structure of Turkish may be confusing for readers with different native languages, we suggest reading [29] and [30] for further details. In this section, we concisely explained our methodology for classifying a document as fake or valid. In the next section, we provide test results and evaluate the findings.

## 4. Results and Evaluation

In the previous section, we have explained how we constructed 4 different lexicons and how these models are used for fake news detection in Turkish. In this section, we outline the results of our experiments to demonstrate the performance of each model and present a comparative analysis. We would like to remind that our task is to label potential fake/fabricated news for facilitating further examination. Hence, we focus on minimizing the amount of fake news which were accidentally labeled as valid news. In the rest of this section, we provide several statistics computed from the labelling of documents residing in the test datasets. Note that, these documents have not been introduced to the training phase where we constructed the lexicons. So, it is possible that there exist terms in these documents which do not exist in the lexicons. If we observe such a word in a document, we assume a term score of 0 for both fake and valid scores of that term.

Initially, we wanted to create a statistically sound benchmark for our testing. Hence, we used the training datasets $D_{tr}^F$ and $D_{tr}^V$ with 5-fold cross-validation to establish test statistics within the training datasets. Table 10 outlines the results of these tests by providing the mean scores for each lexicon.

**Table 10.** 5-fold cross-validation results on the training dataset.

| Model | Precision | Recall | Accuracy | F1 Score |
|---|---|---|---|---|
| Raw | 79.0 | 90.0 | 83.04 | 84.12 |
| Root/Stem | 76.02 | 86.8 | 79.5 | 80.77 |
| Raw+Pos | 75.43 | 81.58 | 77.5 | 78.37 |
| Suffixes | 74.15 | 75.16 | 74.5 | 74.64 |

As seen from this table, considering only the training datasets the best classification accuracy is obtained with the RAW Form model at 83.04. The ROOT model takes the second spot, whereas the RAW+POS model takes the third. SUFFIX model is found to be the least successful model during the validation phase with a classification accuracy of 74.5.

Next, we use the whole of the training datasets to train our models and test with the test datasets $D_{ts}^V$ and $D_{ts}^F$. Hence, 210 valid and 210 fake documents were tested. We provide confusion matrices as well as error statistics for each model separately in Tables 11, 12, 13, 14.

**Table 11.** Confusion matrix and error statistics for the RAW Form Model.

| Prediction | Actual | |
|---|---|---|
| | Fake | Valid |
| Fake | 195 | 48 |
| Valid | 15 | 162 |

| Precision | Recall | Accuracy | F1 Score |
|---|---|---|---|
| 0,802 | 0,929 | 0,85 | |

Table 11 shows the confusion matrix for the RAW Form Model. Out of the 210 FAKE test cases, the model was able to label 195 of them as FAKE, and 15 was erroneously labeled as VALID news. This results in a recall value of 0.929. The precision is lower than recall, however still acceptably high at a value of 0.802. Overall accuracy is 0.85. Successfully identifying about %93 of fake news without generating an unacceptable amount false positive, the RAW Form Model becomes the most promising among the four.

**Table 12.** Confusion matrix and error statistics for the ROOT Form Model.

| Prediction | Actual | |
|---|---|---|
| | Fake | Valid |
| Fake | 193 | 64 |
| Valid | 17 | 146 |

| Precision | Recall | Accuracy | F1 Score |
|---|---|---|---|
| 0,751 | 0,919 | 0,807 | 0,827 |

Presented in Table 12, ROOT Form Model includes only roots of words and achieves a recall of 0.919, only slightly below RAW Form Model, failing to identify two additional fake news texts. However, a larger number of false positives is also generated, resulting in a lower precision than RAW Form Model. Overall ROOT Form Model becomes the second most successful model and shows that root terms of words carry a lot of information regarding the validity of the text.

Next, we turn to RAW+POS Form Model, whose results are presented in Table 13. Although the RAW+POS Form Model terms carry more information than RAW Form Model terms, both the recall and precision values are worse. Compared to ROOT, RAW+POS models has a worse recall, a slightly better precision, and the same overall accuracy. However, recall being our primary goal, this model ranks third among the four.

**Table 13.** Confusion matrix and error statistics for the RAW+POS Form Model.

| Prediction | Actual | |
|---|---|---|
| | Fake | Valid |
| Fake | 188 | 59 |
| Valid | 22 | 151 |

| Precision | Recall | Accuracy | F1 Score |
|---|---|---|---|
| 0,761 | 0,895 | 0,807 | 0,823 |

**Table 14.** Confusion matrix and error statistics for the SUFFIX Form Model.

| Prediction | Actual | | Precision | Recall | Accuracy | F1 Score |
|---|---|---|---|---|---|---|
| | Fake | Valid | | | | |
| Fake | 171 | 60 | 0,74 | 0,814 | 0,864 | 0,776 |
| Valid | 39 | 150 | | | | |

Lastly, we present the results for the SUFFIX Form Model in Table 14. Although, we achieve the worst scores in every statistic in this model, we still believe that the results are impressive and worth discussing. The most impressive feat here is the fact that SUFFIX Form Model does not contain any information about the actual words used in the text. It simply uses the suffix information to evaluate the text as fake or valid. Considering only the suffix groups used within the text, SUFFIX Form Model achieves an impressive 0.814 recall ratio, corresponding to 171 hits in 210 fakes. Although, this is lower than the other lexicons' recall ratios, SUFFIX Form Model achieves this completely unaware of the content of the text.

Finally, we compare the test results with the results obtained from cross-validation testing within the training datasets. It can be seen by comparing the results in Tables 10, 11, 12, 13, and 14 almost all scores have improved in the final testing. This is an expected result as we are using a larger dataset for training and a larger dataset for testing. The cross-validation statistics we obtained using the training datasets alone provided a statistical benchmark to us. It can be said that the final testing scores are reasonably better than these benchmarks and statistically sound within the frame they draw.

## 5. Discussion

In this paper, we presented the first scholarly known study for fake news detection in Turkish based on lexicons. Our study includes collecting a large data set of labeled (fake/valid) Turkish news texts, generating four different lexicons based on these datasets, and providing a highly successful workflow for evaluating news texts using these models. The models we constructed differs by the terms used to generate them. Using the powerful agglutinative structure of the Turkish language we generate a raw word form, a root words form, a raw word with POS tags form, and a suffix form. Although similar studies with respect to raw words, root words and POS tags have been conducted in the literature for different purposes, a model generated using only the suffixes combinations of words is a unique contribution of our paper to the literature.

Our results show that the model generated using the raw forms of the words is the most successful in detecting fake news. This model achieved a recall ratio of 0.929, without generating an unacceptable level of false positives. The model based on root words, and the model based on raw words paired with part-of-speech tags are also quite successful in detecting fake news. The suffix model came last; however, it achieved a significantly high recall of 0.814. Although, this is lower than the other models, considering that there is absolutely no content awareness in this model, this is an impressive ratio. This is a clear indication that suffixes carry a lot of information in agglutinative languages and should be directly considered in similar studies, such as sentiment analysis.

## 6. Conclusions

In the future, we are expecting to enlarge our training datasets with more labeled fake news, further increasing our recall and precision values. We are planning to experiment with reduced models containing only a very small percentage of the models presented in this study to understand whether a core lexicon can be extracted without compromising the recall and precision values.

Finally, we are expecting to merge all four models to come up with an ensemble model that has a higher recall and precision than all four individual models.

In this study, we focused on lexicon-based fake news detection and showed that it can be very effective in Turkish. However, we note that there are many other alternative methods for fake news detection, including stylometry analysis, lexical diversity analysis, punctuation analysis, n-gram based lexicons etc. It is also possible to merge all these studies under a machine learning model, each individual analysis providing a feature value for the learning algorithm. Through this kind of approach, we conjecture that very high recall and precision values can be achieved.

**Acknowledgement:** The authors would like to thank fact-checking organization ``teyit.org" an organization compliant to the principles of IFCN (International Fact-Checking Network) for sharing the data they own.

The dataset will be publicly available for download at (https://github.com/umlab20/FanLexTR) in the near future.

**Appendix A. All results related to 5-fold cross-validation on the training dataset.**

| Dataset Clusters | Models | Precision | Recall | Accuracy | F1 Score |
|---|---|---|---|---|---|
| Set-1 | Raw | 78.846 | 85,416 | 81,25 | 82 |
| | Root | 75,564 | 83,75 | 78,333 | 79,44 |
| | Raw+Pos | 75,806 | 78,333 | 76,667 | 77,049 |
| | Suff | 73,19 | 71,667 | 72,708 | 72,421 |
| Set-2 | Raw | 79,347 | 91,25 | 83,75 | 84,883 |
| | Root | 75,94 | 84,167 | 78,75 | 79,841 |
| | Raw+Pos | 75,572 | 82,5 | 77,917 | 78,884 |
| | Suff | 74,059 | 73,75 | 73,958 | 73,903 |
| Set-3 | Raw | 78,445 | 92,5 | 83,541 | 84,894 |
| | Root | 75,357 | 87,917 | 79,583 | 81,153 |
| | Raw+Pos | 74,532 | 82,917 | 77,292 | 78,501 |
| | Suff | 74,089 | 76,25 | 74,792 | 75,154 |
| Set-4 | Raw | 79,71 | 91,667 | 84,167 | 85,271 |
| | Root | 75,547 | 86,25 | 79,167 | 80,544 |
| | Raw+Pos | 75 | 81,25 | 78,083 | 78 |
| | Suff | 74,477 | 74,167 | 74,375 | 74,322 |
| Set-5 | Raw | 78,677 | 89,167 | 82,5 | 83,593 |
| | Root | 77,737 | 88,75 | 81,667 | 82,88 |
| | Raw+Pos | 76,245 | 82,917 | 78,541 | 79,441 |
| | Suff | 75,294 | 80 | 76,875 | 77,575 |